\pdfoutput=1
\documentclass[11pt]{article}

\usepackage[]{EMNLP2022}

\usepackage{times}
\usepackage{latexsym}

\usepackage[english]{babel} % To obtain English text with the blindtext package
\usepackage{blindtext}

\usepackage{enumitem}
    \setlist[description]{nosep,topsep=0pt}

\usepackage[T1]{fontenc}
\usepackage[utf8]{inputenc}

\usepackage{microtype}

\usepackage{inconsolata}

\usepackage{graphicx}
\usepackage{makecell}
\definecolor{ForestGreen}{rgb}{0.13, 0.55, 0.13}

\newcommand{\red}[1]{\textcolor{red}{#1}}

\newcommand{\green}[1]{\textcolor{ForestGreen}{#1}}
\usepackage[ruled,vlined]{algorithm2e}

\newcommand{\model}{\emph{DREAM-FLUTE}}
\newcommand{\eat}[1]{}

\title{Just-DREAM-about-it: Figurative Language Understanding with \model{}}
\author{ Yuling Gu, Yao Fu, Valentina Pyatkin, Ian Magnusson,\\
\textbf{Bhavana Dalvi Mishra, Peter Clark}\\
Allen Institute for AI, Seattle, WA \\
\texttt{yulingg@allenai.org}
}
\begin{document}
\maketitle
\begin{abstract}
\eat{
% OPTION 2 (better if true): the first -- no idea if it is the first given that I don't know what the other groups tried... 
Figurative language (e.g., ``he flew like the wind'') is challenging to understand as much of the information is conveyed implicitly. We hypothesize that to perform this task well, the reader needs to mentally elaborate the scene being described to identify the appropriate meaning. Here we present \model, a figurative language understanding system that first forms a ``mental model'' of situations described in the premise and hypothesis before making an entailment/contradiction decision and generating an explanation. It internally uses an existing scene elaboration model, DREAM, for constructing its ``mental model.'' In the FigLang2022 Shared Task evaluation, \model{} achieved (joint) first place (Acc@60=63.3\%), and can perform even better with ensemble techniques, demonstrating the effectiveness of this approach.\footnote{We make our code and models publicly available at [to-be-released-URL-if-accepted].} More generally, this work suggests that adding a reflective component to pretrained language models can improve their performance beyond standard fine-tuning (3.3\%   improvement in Acc@60).
}

% Slightly modified version - Pete
Figurative language (e.g., ``he flew like the wind'') is challenging to understand, as it is hard to tell what implicit information is being conveyed from the surface form alone. We hypothesize that to perform this task well, the reader needs to mentally elaborate the scene being described to identify a sensible meaning of the language. We present \model, a figurative language understanding system that does this, first forming a ``mental model'' of situations described in a premise and hypothesis before making an entailment/contradiction decision and generating an explanation. \model{} uses an existing scene elaboration model, DREAM, for constructing its ``mental model.'' In the FigLang2022 Shared Task evaluation, \model{} achieved (joint) first place (Acc@60=63.3\%), and can perform even better with ensemble techniques, demonstrating the effectiveness of this approach.\footnote{We make our code and models publicly available at \url{https://github.com/allenai/dream}.} More generally, this work suggests that adding a reflective component to pretrained language models can improve their performance beyond standard fine-tuning (3.3\%   improvement in Acc@60).

%This paper details the figurative language understanding system we submitted at the FigLang2022 Shared Task (A Shared Task on Understanding Figurative Language). The task involves labeling literal, figurative sentence pairs with entailment/contradiction labels and providing the associated explanations. Our single T5-3B based model using scene elaboration (on likely consequence of events in premise and hypothesis as additional context) improves explanation quality compared to a model without such elaborations and significantly outperforms the T5-3B resource-rich baseline provided by the organizers. Our system was ranked first on the leaderboard, with Acc@0 = 94.7\%,  Acc@50 = 88.9\%, and Acc@60 = 63.3\% on the test set. On top of this, we demonstrate that with a further use of additional context, an ensemble system can achieve even better performance:  Acc@0 = 95.9\%,  Acc@50 = 89.8\%, and Acc@60 = 63.7\%.\footnote{We make our code and models publicly available at [to-be-released-URL-if-accepted].}
\end{abstract}

\section{Introduction}
\label{sec:intro}
\begin{figure}[t]
\centering
     \includegraphics[width=0.9\columnwidth]{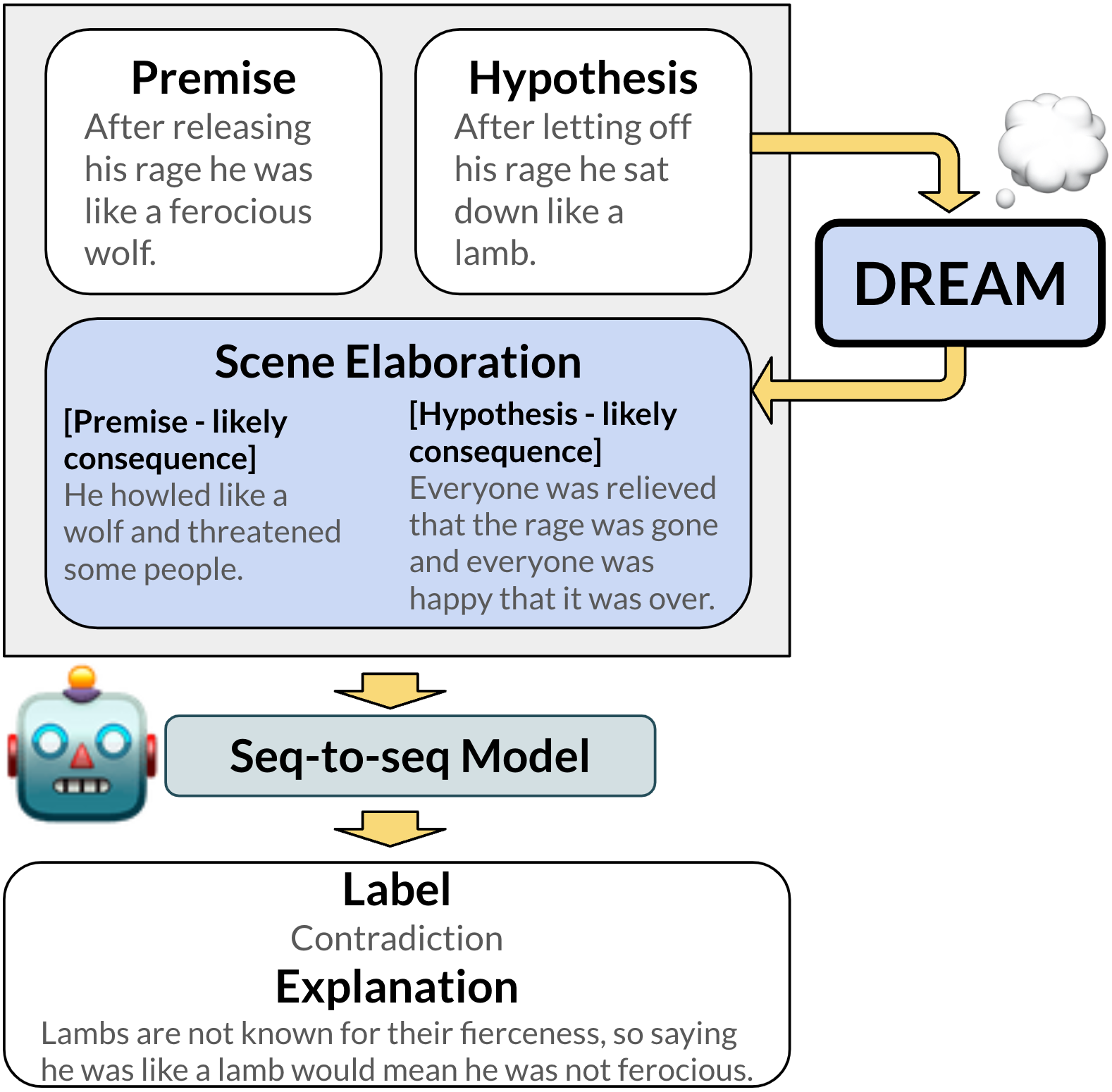}	   
\caption{Overview of \model{}:  It first uses DREAM \citep{gu-etal-2022-dream} to generate an elaboration of the situation in the premise and hypothesis (separately), then uses this additional context for entailment classification and explanation generation. 
\model{} (consequence), using the ``likely consequence''  elaboration dimension as additional context, achieved top scores. Such systems also form the building blocks of \model{} (ensemble), our best system. 
\label{fig:overview}}

\end{figure}
% Edit figures here: https://docs.google.com/presentation/d/1jS4CV4rJDePAy7IL8Kvqakgoh9FBm89xJx7BVUEs_eI/edit?usp=sharing

Understanding figurative language is a particularly challenging problem in NLP since the underlying meaning of the utterance is very different from the surface meaning of its constituent words \citep{stowe-etal-2022-impli}. In this paper we focus on the task of recognizing and explaining textual entailment between a premise and hypothesis involving figurative language (FigLang 2022 Shared Task in~\citealp{Chakrabarty2022FLUTE}). %We believe that to perform this task well, a model needs to mentally elaborate the scene being described to identify sensible meanings of the language. To investigate this further, 
We propose \model{},\footnote{Using DREAM \citep{gu-etal-2022-dream} on FLUTE: Figurative Language Understanding through Textual Explanations \cite{Chakrabarty2022FLUTE}.} a system that makes use of scene elaboration for building a ``mental model'' of the situations presented in the premise and hypothesis to detect textual entailment between them (see Figure \ref{fig:overview}).

%In this paper, we describe our approach to tackling the FigLang 2022 shared task on figurative language understanding \citep{Chakrabarty2022FLUTE}. 
The design of \model{} builds upon the scene elaboration model, DREAM, presented by \citet{gu-etal-2022-dream}. DREAM uses a T5-based \citep{raffel-JMLR-t5} sequence-to-sequence model to generate additional, pertinent details about each given situation in the input text, along key conceptual dimensions informed by cognitive science, story understanding and planning literature \citep{Minsky1974AFF, Dyer1983TheRO, mueller1985daydreaming, mueller1990daydreaming}. Using such scene elaboration as additional context has been shown to improve question-answering (QA) performance on different models and across different downstream tasks %involving situational QA 
such as ETHICS \citep{hendrycks2020aligning}, CODAH \citep{alisa2019codah} and Social IQA \citep{sap2019socialiqa}.

\begin{figure*}[t]
\centering
     \includegraphics[width=0.8\linewidth]{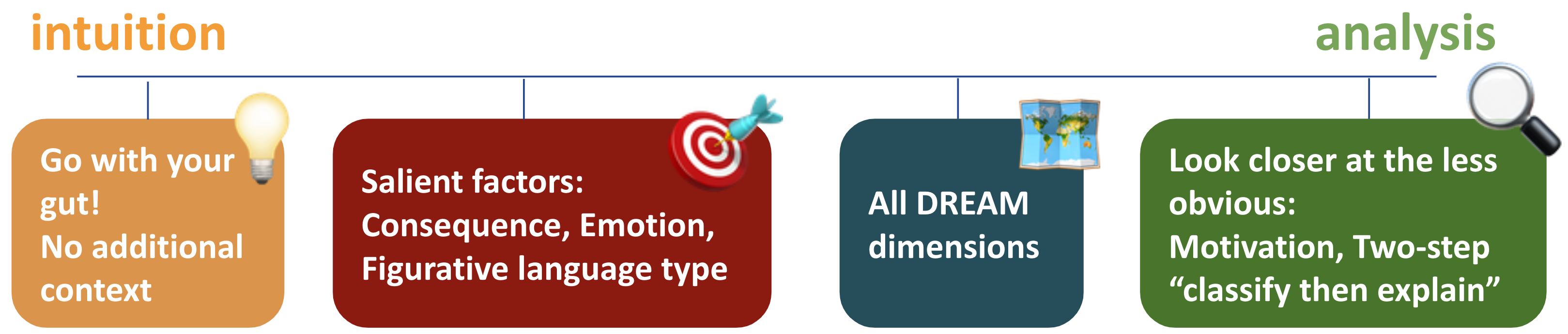}   
\caption{A cognitive continuum implemented to account for different levels of intuition and analysis. \label{fig:continuum}}
\end{figure*}

To adapt it for the figurative language understanding shared task, we made three significant extensions to using DREAM that have not been previously explored.
% in \citet{gu-etal-2022-dream}. 
First, we incorporate DREAM for elaborating the premise and hypothesis in a natural language inference (NLI) task involving figurative language understanding \citep{chakrabarty-etal-2021-figurative, stowe-etal-2022-impli}. We hypothesize that such additional, pertinent details could also improve a model's ability to judge whether there is an entailment or contradiction between the premise and hypothesis. This could be especially helpful for the instances that use figurative language, where the underlying meaning might be opaque to the model: further elaborating the context can make certain inferences more explicit. Second, beyond improvements on label prediction accuracy (i.e. choosing from multiple-choice options) shown in \citet{gu-etal-2022-dream}, our work uncovers the use of such additional context for improving explanation quality. And lastly, we exploit the dimensions in DREAM to train different models for an ensemble system representing a cognitive continuum (Figure \ref{fig:continuum}), further improving accuracy and explanation quality.

Our approach is easily adaptable to other language models, and task-agnostic in format (e.g. QA or NLI) and domain (e.g. ethical decisions or figurative language understanding). We demonstrate the effectiveness of our single model system in terms of achieving top scores in the task, as well as the flexibility of implementing an ensemble system that not only yields further improvements for this task but also allows customization to suit the requirements of different downstream applications. 

\section{Approach}
\label{sec:system}
We first describe our single model systems in Section~\ref{subsec:single_model}. 
Next, we present a two-step ``classify then explain'' pipeline in Section~\ref{subsec:two_step}.
In Section~\ref{subsec:ensemble},
we take advantage of all information learned by the different models and
propose an ensemble approach inspired by cognitive science.

% insert picture of line
% Edit figures here: https://docs.google.com/presentation/d/1jS4CV4rJDePAy7IL8Kvqakgoh9FBm89xJx7BVUEs_eI/edit?usp=sharing

\subsection{Single Model Systems} \label{subsec:single_model}

% \subsubsection{Problem Statement} \label{subsec:input-output-design}

Given an input <Premise, Hypothesis> sentence pair, 
the task has two goals: 
(1). first classify the relationship between the premise and hypothesis (\textit{entailment} or \textit{contradiction});
then (2). generate a textual explanation about why the premise entails/contradicts the hypothesis. 
Figure~\ref{fig:overview} shows an example.
We further consider two additional pieces of information for performance improvements: 
(1). the type of the figurative language (\textit{simile}, \textit{metaphor}, \textit{sarcasm}, \textit{idiom}, and \textit{creative paraphrase}) which is provided in the training data (but not the test data);
(2). the elaboration of situations in the premise-hypothesis pair provided by DREAM, which gives additional information about the \textit{consequence}, \textit{emotion}, \textit{motivation}, or \textit{social norm} of the input.
In Appendix \ref{sec:ex}, we provide intuitive examples showing why such additional information could help this figurative language task.
% Intuitively, these information are helpful to this task because they [Need a one-sentence intuitive explanation about why DREAM helps performance]
%We will show that utilizing information provided by DREAM is our key to performance improvement.
%We will show that utilizing such information can improve performance.

% For this figurative language understanding NLI task, the given inputs are <Premise, Hypothesis> sentence pairs, and the outputs to be evaluated are the ``Entailment'' or ``Contradiction" labels and associated explanations with respect to the meaning of the figurative language expression(s) in the input. An example is shown in Figure \ref{fig:overview}.
% Through our submissions, for single model systems, we evaluated the effect of jointly predicting the type of figurative language as output, as well as providing additional elaboration of situations in the premise and hypothesis along different dimensions (e.g. consequence, emotion, motivation, social norm).
% \footnote{In Appendix \ref{sec:ex}, we  discuss how these designs were motivated by observations from examining some examples.}

% \paragraph{System 1: Using original data}
\noindent \textbf{System 1: Using original data}\quad\quad
Given the  <Premise, Hypothesis, Label, Explanation> in the original data, we first trained a sequence-to-sequence model for the figurative language task using the following input-output format:
\begin{description}
   \item[Input] <Premise> <Hypothesis>
   \item[Output] <Label> <Explanation>
\end{description}
% \noindent \textbf{Input:} Premise: <Premise> Hypothesis: <Hypothesis> Is there a contradiction or entailment between the premise and hypothesis?

% \noindent \textbf{Output:} Answer : <Label> Explanation : <Explanation>

%\noindent This involves making use of just the original data, and inserting the question "Is there a contradiction or entailment between the premise and hypothesis?" to prompt the model for the NLI task.

% \paragraph{System 2: Jointly predicting the type of figurative language}
\noindent \textbf{System 2: Jointly predicting the type of figurative language}\quad\quad
Using type of figurative language provided as part of the training set \citep{Chakrabarty2022FLUTE}, one of our models jointly predicts the type of figurative language, together with the target label and explanation:
\begin{description}
   \item[Input] <Premise> <Hypothesis> 
   \item[Output] <Figurative-Language-Type> <Label> <Explanation>
\end{description}
% \noindent \textbf{Input:} Premise: <Premise> Hypothesis: <Hypothesis> What is the type of figurative language involved? Is there a contradiction or entailment between the premise and hypothesis?
% \noindent \textbf{Output:} Answer : [Type] <Figurative-Language-Type> [Label] <Label> Explanation : <Explanation>
%This could potentially guide the explanations based on predicted type of figurative language.

% \input{EMNLP 2022/011_model_more_index}

% \paragraph{Systems 3, 4, 5, 6, and 7: Providing DREAM's different dimensions as input context}
\noindent \textbf{Systems 3: \model{} - Providing DREAM's different dimensions as input context}\quad \quad
We adapt DREAM's scene elaborations \citep{gu-etal-2022-dream} for the figurative language understanding NLI task by using the DREAM model to generate elaborations for the premise and hypothesis separately. This allows us to investigate if similarities or differences in the scene elaborations for the premise and hypothesis will provide useful signals for entailment/contradiction label prediction and improving explanation quality. Figure \ref{fig:overview} gives an overview of such systems and the input-output format is:
% System 3 is designed to incorporate the emotion dimension of DREAM's scene elaborations:
\begin{description}
   \item[Input] <Premise> <Premise-elaboration-from-DREAM> <Hypothesis> <Hypothesis-elaboration-from-DREAM>
   \item[Output] <Label> <Explanation>
   %\item[Output] Answer : <Label> Explanation : <Explanation>
\end{description}
% \noindent \textbf{Input:} Premise: <Premise> <Premise-emotion-from-DREAM> Hypothesis: <Hypothesis> <Hypothesis-emotion-from-DREAM> What is the type of figurative language involved? Is there a contradiction or entailment between the premise and hypothesis? 
%\noindent \textbf{Output:} Answer : <Label> Explanation : <Explanation>
where the scene elaboration dimensions from DREAM are: 
\textit{consequence},
\textit{emotion}, \textit{motivation}, and \textit{social norm}.
We also consider a system incorporating all these dimensions as additional context.
%The output format is the same as that for System 1.
% Systems 4, 5 and 6 differ from System 3 in that they have inputs enriched with DREAM-generated motivation, likely consequence and social norm respectively.  While these systems each incorporates a different scene elaboration dimension as additional context, System 7 incorporates all dimensions provided by dream.\footnote{In our setup, the dimensions were presented in the following order: social norm, emotion, motivation, and likely consequence. We leave it for future work to explore different ordering of dimensions and the impact of that on various task performance. For an idea of how ordering may matter, please refer to discussion in Section \ref{subsec:ensemble} where order was taken into account in designing the ensemble system.} 

\subsection{Two-step System: Classify then explain} \label{subsec:two_step}
In contrast to Systems 1 to 3 where the entailment/contradiction label and associated explanation are predicted jointly, System 4 uses a two-step ``classify then explain'' pipeline. Previous work on generating explanations have discussed the difference between predicting and generating respective rationalizations in a pipeline vs. jointly. \citet{wiegreffe2021measuring} showed that for reasoning tasks pipelines work less well than models which jointly predict and explain. \citet{hase2020leakage} compared rationalizing methods (first predict label and then the explanation) to reasoning methods (predict the explanation first), and showed that rationalization methods perform better.
It is therefore of interest to compare such different approaches for explanation generation also for the figurative language task.
%\todo{cite some paper where generating explanation together with label is harder to motivate this - Valentina}

\subsection{Ensemble System: A cognitive continuum} \label{subsec:ensemble}

We take advantage of ensembling to use information learned by Systems 1 to 4 together in \model{} (ensemble). 
For entailment/contradiction label prediction, 
the top 5 system variants were chosen based on validation Acc@0 (Table \ref{results-table} \textit{\green{green italicized}}) scores, 
and used for majority voting.

\citet{brachman2022machines} note that several psychologists claim 
``there is a \textit{cognitive continuum} between endpoints that they call \textit{intuition} and \textit{analysis}.''
Likewise, in rationalizing, our different system variants can be viewed as different points on this continuum.
For generating explanations, 
Systems 1 to 4 were used as building blocks for \model{} (ensemble)
(excluding the model with social norm due to its low scores on the validation set)
% \footnote{System 6 was excluded due to its low scores on the validation partition.} 
to implement such a continuum that includes various levels of intuition and analysis (Figure \ref{fig:continuum}).
Specifically, given the entailment label from majority voting, the ensemble looks for the first of the ordered models that agrees with the ensemble label, then uses its explanation. 

Our approach first considers more salient factors (Systems 2, 3 (consequence, emotion)) which can inform the content and style of explanation: likely consequence of the actions and the emotions of characters, which can possibly tease apart whether the sentence pairs entail/contradict,\footnote{E.g. If one situation involves an action leading to good outcome whereas another leads to bad outcome, that is a clear sign (that gives you strong intuition) for contradiction. Whereas, if the premise and hypothesis both describe situations where a person would be happy, that provides intuition for entailment. See Table \ref{analysis-dream-dimensions-2} for examples from task data.} as well as type of figurative language which can inform the style of explanation.\footnote{See Appendix \ref{sec:ex} and Table \ref{type-infl-expl}.}
%E.g. If the person’s emotion is scared in one case and fearless in another, that’s a clear sign (that gives you strong intuition) for contradiction.
Next, we take a step back and look at the bigger picture, in considering all DREAM dimensions \citep{gu-etal-2022-dream} (System 3 (all dimensions)). Then we examine some of the less salient dimensions more closely (Systems 3 (motivation), 4). And finally, we use the explanation in the case when there is no context at all (System 1). %Further details are in Appendix \ref{sec:ensemble-alg}.
More details about this ordering and the pseudocode for ensembling can be found in Appendix \ref{sec:ensemble-alg}.

\section{Experiment Settings}
\label{sec:experiment}
% \subsubsection{Training Details}
% \paragraph{Data} 
\noindent \textbf{Data} \quad\quad
This shared task has a two-phases timeline: the development phase then the test phase. 
During the development phase, 
%(before August 15th 2022)
$\sim$7500 samples are provided as the training set.
% for the shared task \citep{Chakrabarty2022FLUTE}. 
We used a 80-20 split
% \footnote{In creating the split, the provided data was shuffled using a random seed of 42.} 
to create our own training (6027 samples) and validation (1507 samples) partitions on which we build our models. 
% All models were built based these training and validation sets. 
Later at the test phase, 
%(after August 15th 2022)
separate 1500 test samples (without gold labels) are released on which all models are tested. 
Note that our model is primarily developed during the training phase without having access to the test data.

\begin{table*}
\centering
\small
% \normalsize
\begin{tabular}{@{}ll|ccc|ccc@{}}
\hline
%  \multirow{2}{*}{\multicolumn{2}{c|}{\textbf{System}}}  & \multicolumn{3}{c|}{\textbf{Our validation partition}} & \multicolumn{3}{c}{\textbf{Official test partition}}\\
 \multicolumn{2}{c|}{\textbf{System}}  & \multicolumn{3}{c|}{\textbf{Our validation partition}} & \multicolumn{3}{c}{\textbf{Official test partition}}\\
{} & {}  & \textbf{Acc@0} &  \textbf{Acc@50} &  \textbf{Acc@60} & \textbf{Acc@0} &  \textbf{Acc@50} &  \textbf{Acc@60}\\
\hline
  & T5-3B (official baseline)  &  -- &  --  &  --  &  76.7    &    69.1   &    44.3\\
%   & Prompting GPT3  &  85.0 &  75.9  &  47.3  &  --    &    --   &    -- \\
\hline
1 & Original data &  \textit{\green{94.8}} &  89.0  &  66.9  &  94.7    &    88.7    &    60.4\\
2 & + Figurative language type &  \textit{\green{94.9}} &  89.8  &  66.5    &    94.6    &    87.8    &    61.3\\
% 3 & + DREAM explanations  &   &    &  &   &     & \\   
3 & \model{}  &   &    &  &   &     & \\  
 & \quad emotion &  94.2 &  89.3  &  65.0  &  93.9    &    88.3   &     61.7\\
 & \quad  motivation &  \textit{\green{95.4}} &  90.2  &  66.2  &  94.5  &  87.7   &    60.3\\
 & \quad  consequence &  94.3 &  90.1  &  65.8  &  94.7    &    88.9    &    \textbf{63.3}\\
 & \quad  social norm &  93.1 &  88.3  &  64.2  &  92.3 & 86.4    &    60.6\\
 & \quad  all 4 dimensions &  \textit{\green{95.2}} &  89.4  &  66.6  &  94.3   &     87.7   &    60.0\\
4 & Classify then explain &  \textit{\green{95.0}} &  90.5  &  66.6  &  95.1   &     89.4   &     61.1\\
% 5 & Ensemble (of continuum) &  \textbf{96.4} &  \textbf{92.1}  &  \textbf{67.0}  &  \textbf{95.9}  &    \textbf{89.8}  &  \textbf{63.7}\\
5 & \model{} (ensemble) &  \textbf{96.4} &  \textbf{92.1}  &  \textbf{67.0}  &  \textbf{95.9}  &    \textbf{89.8}  &  \textbf{63.7}\\
\hline
\end{tabular}
\caption{\label{results-table}
Results on our validation set and the official test set. Amongst the non-ensemble methods, System 3 with likely consequence, i.e. \model{} (consequence), performed the best on the test set in terms of Acc@60 which was used for ranking submissions on the leaderboard. This system was already ranked first, but further gains can still be achieved using ensembling in System 5,  \model{} (ensemble). \textit{\green{Green italics}} indicates systems selected for label prediction in the ensemble system, using validation Acc@0.
}
\end{table*}

% \paragraph{Model}
\noindent \textbf{Model} \quad\quad
We train all models with a 
T5-3B backbone using the data formats detailed in Section~\ref{subsec:single_model}. 
The size of the model is the same as the officially provided fine-tuned T5 baseline.
We use the Huggingface implementation \citep{Wolf2019HuggingFacesTS, wolf-etal-2020-transformers},
% \footnote{\url{https://github.com/huggingface/transformers}} 
based on PyTorch \citep{Paszke2019pytorch}. 
For each system, we fine-tune the 3B version of T5 \citep{raffel-JMLR-t5} for 3 epochs using an Adam Optimizer and a learning rate of 5e-05, selecting the best checkpoint based on the lowest validation loss.
A more detailed list of hyperparameters used can be found in Appendix \ref{sec:hyper-param}.
% \footnote{A more detailed list of hyperparameters used can be found in Appendix \ref{sec:hyper-param}.}

% \paragraph{Baselines} 
% \noindent \textbf{Baselines} \quad\quad
% We consider two baselines: (1) the fine-tuned resource rich T5 model provided by the workshop organizer; (2) prompting GPT3 with 8 sampled training instances and each instance is formated as: [premise-hypothesis-explanation-label]

% \paragraph{Evaluation}
\noindent \textbf{Evaluation} \quad\quad
There are two major evaluation metrics:
(1). \textit{accuracy}, which measures if predicted NLI labels are correct; 
(2). \textit{explanation score}, which  measures if  generated explanations are of high quality.
The explanation score is computed as the average of BERTScore~\citep{ZhangKWWA20-bertscore} and BLEURT~\citep{sellam-etal-2020-bleurt} 
on the generated explanation against given references. 
The overall performance metric, Acc@$s$ (Table~\ref{results-table}), 
is a combination of \textit{accuracy} and \textit{explanation score}
where a prediction (label and explanation) counts as correct only when: 
(a) the label is correct, 
and (b) the explantion score is at least $s$ (where $s = 0, 50$ and $60$).
On the official leaderboard, all models are ranked according to Acc@60.

\section{Results and Discussion}

% Table \ref{results-table} shows the performance of our systems.\footnote{We also show the baseline model provided by organizers. For a fair comparison, all our approaches also use the T5-3B model.}
%on the validation and official test partitions are summarized in Table \ref{results-table}. We used our validation partition to guide choices during model training and ensembling. Systems were ranked on the leaderboard based on the official test set.
% insert results table 

% \input{031_tab_results}

\begin{table*}[]
\centering
\tabcolsep=0.11cm
\begin{tabular}{|lll|l|ll|}
\hline
Premise & Hypothesis & \makecell[l]{Gold\\label} &  \makecell[l]{T5's answer \\ + explanation} & \makecell[l]{DREAM\\context} &  \makecell[l]{\model's \\ answer  + explanation}\\
\hline
\makecell[l]{My decision-\\making skills \\are not purely\\ based on \\emotions \\and gut.} & 
\makecell[l]{My personal \\feelings \\color my \\judgment \\in this case.} & (C) & 
\makecell[l]{\red{(E)}: My personal \\feelings color \\my judgment in \\this case means \\that I am \\influenced \\by my emotions \\and gut when \\making decisions.}&
\makecell[l]{\textbf{[Consequence]} \\\textbf{Pre:} I make \\more balanced \\and informed \\decisions.		\\\textbf{Hypo:} I make \\a decision that \\is not in\\ the best interests\\ of the company.\\	}&
\makecell[l]{\green{(C)}: To have personal \\feelings color \\one's judgment \\means to make \\decisions based \\on them, but \\this context describes \\making decisions \\ based on facts and \\ not emotions.}\\
\hline
\makecell[l]{We laid \\in the \\field of \\green grass \\and relaxed.} & 
\makecell[l]{We laid \\in fields \\of gold.} & (E) & 
\makecell[l]{\red{(C)}: Gold is a type\\ of metal that is used \\to make jewelry,\\ while green grass\\ is a type of grass that \\is used for lawns\\ and gardens.}&
\makecell[l]{\textbf{[Emotion]} \\\textbf{Pre:} I (myself)'s \\emotion is happy.\\	\textbf{Hypo:} I (myself)'s \\emotion is happy.}&
\makecell[l]{\green{(E)}: Gold is a \\color that is associated\\ with happiness, \\so the fields\\ of gold are\\ associated with\\ happiness.}\\
\hline

% \makecell[l]{After years \\of service \& \\contribution \\to the \\company, \\he was \\finally\\ promoted.} & 
% \makecell[l]{The company \\released him \\after many \\years of \\service.} & (C) & 
% \makecell[l]{\red{(E)}: The company \\released the \\employee after \\many years of \\service.}&
% \makecell[l]{\textbf{[Motivation]} \\\textbf{Pre:} Company's\\ motivation is to \\recognize his \\hard work.	\\\textbf{Hypo:} Company's \\motivation is \\to get someone \\else to work.\\	}&
% \makecell[l]{\green{(C)}:To release \\someone \\means to let them \\go from a position, \\while to promote \\someone means to \\give them a \\higher position.}\\
% \hline

\end{tabular}
\caption{\label{analysis-dream-dimensions-2}
Examples from the validation set where DREAM generated context consisting of consequence and emotion  helped our proposed model \model{} (System 3) in figurative language understanding. For both the examples a T5-based model that did not have access to additional context (System 1) gave wrong label prediction. DREAM context helped improve both answer accuracy and explanation quality. Labels: (E), (C) refer to Entailment, Contradiction respectively. (Appendix \ref{analysis-dream-context} presents examples where motivation, social norm helped \model.)
}
\end{table*}

\subsection{Better explanation quality}
%Incorporating information from jointly predicting the type of figurative language, as well as using various DREAM dimensions as additional context, we achieved various amounts of improvements in explanation quality compared to the setup with just the original data. 
Table \ref{results-table} shows the performance of our systems.
%For a fair comparison, all our approaches also use the T5-3B model. %except GPT3.
Based on test Acc@60, the following strategies improve explanation quality compared to the setup with just the original data: predicting figurative language type, using emotion, likely consequence, social norm, two-step ``classify then explain'' pipeline, and ensembling. Each non-ensemble system can be seen as guiding the model to focus on a particular direction when reasoning about the entailment/contradiction relationship between a sentence pair. Table \ref{analysis-dream-dimensions-2} and Appendix \ref{analysis-dream-context} 
%Table \ref{analysis-dream-dimensions} 
present examples of how each DREAM dimension helps uncover implicit meaning in the input. \model{} (consequence), by incorporating the likely consequence scene elaboration from DREAM, was already ranked first based on test Acc@60,\footnote{See results of shared task at \url{https://codalab.lisn.upsaclay.fr/competitions/5908\#results}.}
which requires explanations to be of high quality. Figure \ref{fig:overview} shows another example of how elaborating along this dimension can be useful. 
On top of that, \model{} (ensemble), an ensemble system that makes further use of context achieves further improvements (Acc@60 = 63.7\%). The ensemble approach allows for considering these different directions and rationalizing with varying levels of intuition and analysis, then choosing one that fits the current sentence pair, potentially boosting explanation quality.

% copy over example from hackathon slides
%\todo{1 example each of emotion, motivation, consequence, social norm helps wrt system 1- Bhavana}

\subsection{Better label prediction accuracy}
This ensemble system is also our best submission overall with Acc@0 = 95.9\%,  Acc@50 = 89.8\%, where Acc@0 is equivalent to computing label accuracy alone. The better label prediction accuracy could be attributed to using the different individual systems for majority voting, which mimics arriving at a decision by considering different perspectives, ultimately leading to a more well-thought decision.
%\todo{some analysis of w/o vs w dream context}

\subsection{Effect of DREAM generated context} %\label{analysis-dream-context}
%Here we qualitatively analyze how DREAM-generated context for premise and hypothesis helps in the figurative language understanding task. 
We qualitatively analyze how DREAM-generated context 
%for premise and hypothesis 
helps in the figurative language understanding task. 
Table \ref{analysis-dream-dimensions-2} presents examples from our validation set for DREAM dimensions ``consequence'' and ``emotion'' %where such additional context helped correct label prediction and improve explanation. By 
comparing predictions from System 1 (trained using just original data) with those from System 3 (\model{}, which uses scene elaboration from DREAM). These examples illustrate that
similarities and differences along the scene elaboration dimensions provide useful signals to guide entailment/contradiction label prediction and improve explanation quality.

\subsection{More flexibility beyond FigLang2022}
The day-to-day mental activities of humans take place on different parts of the cognitive continuum \citep{brachman2022machines}. DREAM’s scene elaborations give us the different building blocks to implement to such a continuum, and therefore use various levels of intuition and analysis to better come to a decision and rationalize. This approach also allows customization to suit the requirements of different downstream applications, by changing the order of factors to consider on the continuum (e.g. social norm may be more salient for ethical decisions) and considering different pertinent factors (i.e. in place of the figurative language type).

% \section{GPT-3 baseline} \todo{GPT-3 prompting approaches -- Yao}

% 
% Better than GPT-3 with extensive prompt-tuning efforts

% If no finetuning, existing NLI moedels -- a similar approach to improve label accuracy 

\section{Conclusion}
\label{sec:conclustion}
In this work we showed how \model{}, a competitive system for the figurative language understanding NLI task, can be built by utilizing scene elaborations from an existing model, DREAM. Compared to a model without such scene elaborations, \model{} makes use of scene elaboration for building a ``mental model'' of situations in the premise and hypothesis to make inferences more explicit, thus improving label prediction accuracy and explanation quality. \model{} (ensemble) uses different elaborations to form building blocks for implementing a continuum with varying levels of intuition and analysis, modeling deriving answers and rationalizing by considering different positions on a cognitive continuum. This novel use of DREAM not only obtained the highest scores for the figurative language understanding shared task, but could also easily be applied to the situational QA tasks in \citet{gu-etal-2022-dream}, and beyond. Our approach is easily adaptable to other language models, and task-agnostic in format (e.g. QA or NLI) and domain (e.g. ethical decisions or figurative language understanding). More generally, our work demonstrates that adding a reflective component helps to improve answer accuracy and explanation quality in pretrained language models.

\clearpage

%% This section should occur after the conclusion, but before the references. It will not count towards the page limit.  
\section*{Limitations}
Our approach is designed for applications involving natural language understanding for short text (around 1-3 sentences), e.g. in the figurative language NLI task and situational QA tasks tackled in the original DREAM paper. Building on a better understanding for short text, we hope our work can inspire future efforts towards extending the approach for long text too. The current approach presented also requires the use of GPU resources for model training. However, we also demonstrate that using DREAM scene elaboration as additional context yields improvements on label prediction accuracy for an off-the-shelf NLI model, without any training (Table \ref{baseline-mnli-table} in Appendix \ref{mnli-baselines}). 

%% The ethics statement will not count toward the page limit.
\section*{Ethics Statement}
Like any other large-scale
language model, despite the best intentions, there is a risk of our models producing biased or offensive statements as part of the free-form rationalization. We release our models for research purposes only.

\section*{Acknowledgements}
We would like to thank the entire Figurative Language Understanding Shared Task organizing committee for organizing this shared task. 
% Reviewers
We thank the anonymous reviewers for their helpful comments.
% Hackathon
This work was done as part of a Hackathon project during AI2's 2022 Hackathon. We are grateful to the Hackathon organizers, Caitlin Wittlif and Carissa Schoenick, for the great 3-day Hackathon that led to this work.

\bibliography{anthology,custom}
\bibliographystyle{acl_natbib}

\clearpage
\appendix

\section{Examples from training set}
\label{sec:ex}

\begin{table*}[]
\centering
\tabcolsep=0.11cm
\begin{tabular}{|l|l|l|l|l|}
\hline
 \makecell[l]{Type of \\figurative\\ language} & Premise & Hypothesis & Gold label & Gold Explanation\\
\hline
Sarcasm &
\makecell[l]{Yesterday two gangs\\
were fighting just\\
in front of my home.} & 
\makecell[l]{Yesterday I saw\\
two gangs fighting\\
right in front of my\\
house and it totally\\
didn't make me\\
scared at all.} & Contradiction & 
\makecell[l]{The sight of two\\
gangs fighting \textbf{is}\\
\textbf{often} very violent\\
and can invoke fear \\
in people, \textbf{so} someone\\
who saw it and wasn't\\
scared is not\\
being truthful.}\\
\hline
Idiom & 
\makecell[l]{If you want fresh food,\\
just go with your gut\\ 
feeling and you will\\
find villagers happy to\\
sell or trade what\\
they have.} &
\makecell[l]{If you want fresh\\
food, just follow your\\
noses and you will find\\
villagers happy to sell\\
or trade what they have.}& Entailment & 
\makecell[l]{To \textbf{follow your nose}\\
\textbf{means} to trust one's\\
instinct, which is what\\
you would need to do in\\
order to find fresh food.}\\
\hline
\end{tabular}
\caption{\label{type-infl-expl}
Examples from \citet{Chakrabarty2022FLUTE}'s training set. Text in bold illustrate how the style of explanation may depend on the type of figurative language involved.
}
\end{table*}

We randomly sampled around 100 examples from the training set and manually looked at the targeted explanations to get a sense of how explanations for this task look like. We observed that the explanation style may depend on the type of figurative language involved. Table \ref{type-infl-expl} shows some of these examples. For instance, when the type of figurative language is sarcasm, the explanation often starts by describing what is usually the case and then goes into how one of the sentences describes an unusual or unexpected situation. Whereas, if the type is idiom, then the explanation often involves elucidating what the idiom means. This motivated the design of System 2.

Further, we noticed that the gold explanations often involve elements like emotion and motivation of characters. In the first example in Table \ref{type-infl-expl}, for example, identifying the emotions in the premise and hypothesis directly helps us identify the contradiction — in that the person’s emotion is scared in one case and fearless in another. Therefore, we explored elaborating the situations in the given premise and hypothesis along such dimensions using DREAM \citep{gu-etal-2022-dream}. By using DREAM to generate scene elaborations and using that as additional context to the input, we have the different variations of \model{} (System 3). 
\section{Details of input prompt}
\label{sec:prompt}
In training our T5 based sequence-to-sequence models, whenever the target output is the entailment/contradiction label and explanation, we append the question ``Is there a contradiction or entailment between the premise and hypothesis?'' to the input to prompt the model for the NLI task. In the case of System 2, where the model jointly predicts the type of figurative language then the label and explanation, we first append the question ``What is the type of figurative language involved?'' to the input, then append the usual contradiction or entailment question.

\section{Algorithm for ensembling}
\label{sec:ensemble-alg}
The order of systems used in rationalizing when implementing the cognitive continuum described in Section \ref{subsec:ensemble} is as follows: likely consequence, emotion, type of figurative language, all DREAM dimensions, motivation, two-step ``classify then explain,'' no context. Algorithm \ref{alg: ensemble} shows more details on how to obtain the ensemble label and explanation from the individual systems.

Note that beyond the figurative language understanding task, this ensembling approach representing a cognitive continuum could be applied to other tasks, with the possibility of modifying the order of component systems to better suit different applications.

\begin{algorithm}[t]
\small
\DontPrintSemicolon
\SetAlgoLined
\SetKwInput{KwData}{Input}
\SetKwInput{KwResult}{Output}
\KwData{Individual systems' predicted label and explanation} 
\KwResult{Ensemble label; Ensemble explanation}

ensemble\_label $=$ majority\_vote(top5\_Acc@0\_systems\_labels)
ensemble\_explanation $=$ None

\tcp*[l]{ {ordered\_systems takes an order described in Section \ref{sec:ensemble-alg}}}

\For{system\_prediction $\in$ ordered\_systems}{
    \If{system\_prediction.label $==$ ensemble\_label}{
        ensemble\_explanation $=$ system\_prediction.explanation\;
        \textbf{break}
    }
}
 \caption{\small Ensemble - a cognitive continuum}
 \label{alg: ensemble}
\end{algorithm}

\section{Hyperparamters used during training}
\label{sec:hyper-param}

The following hyperparameters were used during training:
\begin{itemize}
  \item learning\_rate: 5e-05
  \item train\_batch\_size: 1
  \item eval\_batch\_size: 1
  \item seed: 42
  \item distributed\_type: multi-GPU
  \item num\_devices: 2
  \item total\_train\_batch\_size: 2
  \item total\_eval\_batch\_size: 2
  \item optimizer: Adam with betas=(0.9,0.999) and epsilon=1e-08
  \item lr\_scheduler\_type: linear
  \item num\_epochs: 3.0
\end{itemize}

\section{Baseline: Off-the-shelf MNLI model} \label{mnli-baselines}
Without any training on the task data, we can similarly achieve better label prediction accuracy if we provide additional context from DREAM as input. Table \ref{baseline-mnli-table} shows that with the off-the-shelf RoBERTa MNLI model \cite{roberta-mnli}, we achieve improvements in accuracy when providing the emotion of characters, and even more improvements if we provide all 4 dimensions generated by DREAM. Since this model is unable produce any explanations, we measure only Acc@0 scores.

\begin{table}[h]
\centering
\small
% \normalsize
\begin{tabular}{|l|ccc|}
\hline
%  \multirow{2}{*}{\multicolumn{2}{c|}{\textbf{System}}}  & \multicolumn{3}{c|}{\textbf{Our validation partition}} & \multicolumn{3}{c}{\textbf{Official test partition}}\\
 \multicolumn{1}{|c|}{\textbf{System}}  & \multicolumn{3}{c|}{\textbf{Our validation partition}} \\
 {}  & \textbf{Acc@0} &  \textbf{Acc@50} &  \textbf{Acc@60} \\
\hline
%   & Prompting GPT3  &  85.0 &  75.9  &  47.3  &  --    &    --   &    -- \\
 RoBERTa MNLI &  73.9  &  --    &    --\\
             \ \ + DREAM emotion &  77.4  &  --    &    --\\
             \ \ + DREAM 4 dimensions &  79.3  &  --    &    --\\
\hline
\model{}(ensemble) &  \textbf{96.4} &  \textbf{92.1}  &  \textbf{67.0} \\
(our model) & & & \\
\hline
\end{tabular}
\caption{\label{baseline-mnli-table}
Comparing off-the-shelf RoBERTa MNLI model \cite{roberta-mnli} to our proposed model on our validation set.
}
\end{table}

\section{Analysis: Effect of DREAM generated context} \label{analysis-dream-context}
Table \ref{analysis-dream-dimensions-remaining-2} presents validation set examples for DREAM dimensions ``motivation'' and ``social norm'' where this additional context helped the model correct its prediction.   

\begin{table*}[]
\centering
\tabcolsep=0.11cm
\begin{tabular}{|lll|l|ll|}
\hline
Premise & Hypothesis & \makecell[l]{Gold \\label} &  \makecell[l]{T5's answer \\ + explanation} & DREAM context &  \makecell[l]{\model's \\ answer  + explanation}\\
\hline
% \makecell[l]{We laid \\in the field \\of green grass \\and relaxed.} & 
% \makecell[l]{We laid \\in fields \\of gold.} & (E) & 
% \makecell[l]{\red{(C)}: Gold is a \\type of metal \\that is used \\to make jewelry,\\ while green grass\\ is a type of \\grass that \\is used for \\lawns and gardens.}&
% \makecell[l]{\textbf{[Emotion]} \\\textbf{Pre:} I (myself)'s \\emotion is happy.\\	\textbf{Hypo:} I (myself)'s \\emotion is happy.}&
% \makecell[l]{\green{(E)}: Gold is a color \\that is associated\\ with happiness, \\so the fields\\ of gold are\\ associated with\\ happiness.}\\
% \hline

\makecell[l]{After years \\of service \\\& contribution \\to the company, \\he was \\finally\\ promoted.} & 
\makecell[l]{The company \\released him \\after many \\years of \\service.} & (C) & 
\makecell[l]{\red{(E)}: The company \\released the \\employee after \\many years of \\service.}&
\makecell[l]{\textbf{[Motivation]} \\\textbf{Pre:} Company's\\ motivation is to \\recognize his \\hard work.	\\\textbf{Hypo:} Company's \\motivation is \\to get someone \\else to work.\\	}&
\makecell[l]{\green{(C)}:To release \\someone \\means to let them \\go from a position, \\while to promote \\someone means to \\give them a \\higher position.}\\
\hline

% \makecell[l]{My decision-\\making skills \\are not purely\\ based on \\emotions \\and gut.} & 
% \makecell[l]{My personal \\feelings \\color my \\judgment \\in this case.} & (C) & 
% \makecell[l]{\red{(E)}: My personal \\feelings color \\my judgment in \\this case means \\that I am \\influenced \\by my emotions \\and gut when \\making decisions.}&
% \makecell[l]{\textbf{[Consequence]} \\\textbf{Pre:} I make \\more balanced \\and informed \\decisions.		\\\textbf{Hypo:} I make \\a decision that \\is not in\\ the best interests\\ of the company.\\	}&
% \makecell[l]{\green{(C)}:To have personal \\feelings color \\one's judgment \\means to make \\decisions based \\on them, but \\this context describes \\making decisions \\ based on facts and \\ not emotions.}\\
% \hline

\makecell[l]{Sure, he\\ snorted just to\\ make me feel\\ even better\\ about the\\ already great\\ situation.} & 
\makecell[l]{Sure, he \\snorted, just\\ rub it in.} & (C) & 
\makecell[l]{\red{(E)}: To rub it in \\means to make \\someone feel \\better about an\\ already great \\situation, \\which is what \\he did by\\ snorting.}&
\makecell[l]{\textbf{[Social Norm]} \\\textbf{Pre:} It's good to \\make people feel\\ better about \\a situation.			\\\textbf{Hypo:} It's rude to \\rub something in \\someone's face\\ when they don't\\ want to.\\	}&
\makecell[l]{\green{(C)}:To rub it in \\means to make \\someone feel bad\\ about themselves, \\but in this sentence\\ he is making the \\speaker feel better \\about the already \\great situation.}\\
\hline

\end{tabular}
\caption{\label{analysis-dream-dimensions-remaining-2}
Examples from the validation set where DREAM generated context consisting of motivation and social norm helped our proposed model \model{} (System 3) in figurative language understanding. For all these examples a T5-based model that did not have access to additional context (System 1) gave wrong label prediction. DREAM context helped improve both answer accuracy and explanation quality. Labels: (E), (C) refer to Entailment, Contradiction respectively. 
}
\end{table*}

% \begin{figure}[t]
% \centering
%      \includegraphics[width=0.9\columnwidth]{figures/roberta-mnli-no-training.png}	   
% \caption{ \label{fig:roberta-mnli-no-training}}

% \end{figure}

\end{document}